\documentclass[nonacm,sigconf]{acmart}

\renewcommand\footnotetextcopyrightpermission[1]{}
\AtBeginDocument{%
  \providecommand\BibTeX{{%
    \normalfont B\kern-0.5em{\scshape i\kern-0.25em b}\kern-0.8em\TeX}}}

\setcopyright{none}
\usepackage[usenames,dvipsnames,svgnames,table]{xcolor}

\usepackage{color}
\usepackage{caption}
\usepackage{subcaption}

\usepackage[most]{tcolorbox}
\usepackage{tikz}
\usepackage{xcolor}

\newtcolorbox{promptbox}[2][]{
    enhanced,
    breakable,
    colback=gray!5,
    colframe=black!60,
    boxrule=0.5pt,
    arc=2pt,
    left=6pt, right=6pt, top=4pt, bottom=4pt,
    fonttitle=\bfseries\small,
    title=#2,
    #1
}

\setcopyright{none}

\begin{document}
\fancyfoot[C]{\thepage}
\title[How Unlikely Is ``Unlikely''?]{How Unlikely Is ``Unlikely''? Assessing Verbal Probability Perception Across Large Language Models}

\author{Christos Petridis}
\email{christos.petridis@temple.edu}
\affiliation{%
  \institution{Department of Computer and Information Sciences \\ Temple University}
  \city{Philadelphia}
  \state{PA}
  \country{USA}
}

\author{Konstantinos Pelechrinis}
\email{kpele@pitt.edu}
\affiliation{%
  \institution{Department of Informatics and Networked Systems \\ University of Pittsburgh}
  \city{Pittsburgh}
  \state{PA}
  \country{USA}
}

\author{Zoran Obradovic}
\email{zoran.obradovic@temple.edu}
\affiliation{%
  \institution{Department of Computer and Information Sciences \\ Temple University}
    \city{Philadelphia}
  \state{PA}
  \country{USA}
}

\renewcommand{\shortauthors}{C. Petridis, K. Pelechrinis, and Z. Obradovic.}

\begin{abstract}
Large language models increasingly produce and interpret verbal probability expressions, yet whether these expressions carry consistent meaning across models (or match human perceptions of uncertainty) remains unknown. We present a systematic cross-model evaluation using a word-to-number mapping task grounded in established human benchmarks. Eleven uncertainty expressions were presented to 19 models under two conditions, forced single-number response and explanation elicitation, alongside a novel bidirectional roundtrip test of internal consistency. LLMs track the human benchmark with surprising fidelity: word ordering is preserved, three anchor points are recovered, and ``possible'' shows the highest variance and cross-model disagreement of any expression tested, consistent with its documented bimodal interpretation in humans. However, models show a systematic upward bias for negative expressions such as ``unlikely'' and ``improbable.'' Explanation elicitation reduces within-model variance while increasing between-model divergence, stabilizing individual models at the cost of inter-model consensus, and the roundtrip experiment reveals clear stratification, with frontier models maintaining coherent bidirectional representations. LLMs thus reproduce the structure of human verbal probability cognition, including its biases, while diverging systematically at the negative end---with implications for any setting where humans and models exchange probabilistic language.

\end{abstract}

\maketitle

\section{Introduction}\label{sec:intro}

Every day, millions of people ask AI systems questions whose answers are uncertain:

\begin{itemize}
    \item Will this treatment work? 
    \item Is this investment risky?
    \item Is this news article trustworthy? 
    \item Is this paper going to be accepted? 
\end{itemize}

As large language models are quickly becoming collaborators in everyday decision making, people rely on them not only to retrieve information but also to evaluate uncertain situations, while weighing alternatives and making recommendations.  
In these interactions, communicating what is known is only part of the challenge, since equally important is communicating how certain that information is. 
In most cases AI systems communicate uncertainty using natural language (``likely'', ``unlikely'',``possible'', ``almost certain'') rather than explicit probabilities \cite{belem2024perceptions,steyvers2025metacognition}. 
However, decades of decision science show that people interpret these expressions differently \cite{budescu1985consistency,wallsten1986measuring,wallsten1993preferences,wintle2019verbal}. 
Different people assign substantially different numerical interpretations to the same words, and those interpretations depend on context, expertise, and prior beliefs. 

The question we want to answer with our work is whether the same is true for generative AI tools and how their interpretations compare to the human ones. 
Without shared interpretations, even perfectly accurate predictions may be misunderstood, leading users to overestimate or underestimate the confidence that an AI system intends to convey. 
Successful human-AI collaboration depends not only on AI making accurate predictions, but also communicating these predictions in a way that they are understandable to the users. 

To investigate this, we conduct a systematic cross-model evaluation of verbal probability perception in large language models. 
We present 11 probability expressions to 19 models drawn from major commercial and open-source families, under two prompt conditions: a forced single-number response and an explanation elicitation condition. 
We ground our analysis in the human benchmark established by Mosteller and Youtz \cite{mosteller1990quantifying}, which aggregates numerical interpretations of probability words across 20 human studies. 
We also extend our analysis with a novel bidirectional roundtrip experiment designed to test the internal consistency of each model's word-to-number mapping. 

Our results indicate that LLMs collectively track the human benchmark with good fidelity. 
In particular, word ordering is preserved across all models and three anchor points (namely,  ``impossible'', ``even chance'', and ``certain'') are recovered with near-zero variance. 
Furthermore, ``possible'' emerges as the single most variable expression across all models, a pattern consistent with the word's documented resistance of stable numerical interpretation in human studies. 
However, LLMs also rate probabilistic expressions as more probable than human norms suggest. 
The three anchor points above form the exception to this pattern.  
We also identify a larger, albeit marginal,  inflation for negative expressions under the forced single number response condition.
Explanation elicitation reduces within-model response variance but also increases divergence across models, while the roundtrip experiment
reveals that while most models
maintain a coherent bidirectional mapping between words and numbers,
a small subset of them show mappings that are statistically indistinguishable
from chance.

These findings have direct implications for human-AI communication. 
LLMs communicate uncertainty in ways that are, to a large extent, consistent with human norms, preserving word ordering, anchoring correctly at the extremes, and reproducing known patterns of ambiguity. 
Their deviations from those norms are real and predictable, concentrated in the non-anchor expressions examined, while the three anchor words (\textit{impossible}, \textit{even chance}, and \textit{certain}) show negligible bias. 
Within the examined expressions, deviation is largest for the inherently ambiguous
\textit{possible}, and we find weak evidence that negatively worded expressions exhibit slightly higher bias under the forced condition.
Understanding where these deviations occur, and how they vary across models and prompt conditions, is a prerequisite for any application where probabilistic language is important.


\section{Related Work}
\label{sec:related}

The interpretation of verbal probability expressions has been studied for over half a century. 
Early work established that people assign widely varying numerical interpretations to the same words, and that those interpretations are stable across populations and contexts. 
Mosteller and Youtz \cite{mosteller1990quantifying} synthesized results from 20 studies covering 52 expressions, finding broad cross-population agreement for most words with one notable exception, namely, the word ``possible''. 
This word showed a distinctly bimodal distribution, with respondents splitting between near-zero and near-fifty interpretations. 
This instability has been consistently identified in subsequent work \cite{wintle2019verbal, hashim2024verbal}. 
Smithson et al. \cite{smithson2012never} formally documented a systematic asymmetry between positively and negatively worded expressions, showing that negative wording reduces precision in numerical translations independently of any shift in the mean response. 
Wintle et al. \cite{wintle2019verbal} replicated this pattern in a larger general-population sample, further showing lower consistency with normative guidelines for negative expressions, and attributed the effect to the greater difficulty of calibrating events that do not occur.
Across studies, five expressions, namely, ``very likely'', ``likely'', ``possible'', ``unlikely'', and ``very unlikely'', show sufficiently overlapping numerical interpretations to form the basis of standardized verbal probability scales in clinical and intelligence communication \cite{hashim2024verbal}.  
However, ``possible'' is still a notable exception to this pattern, and, despite its consistent ordinal placement, it produces the widest distribution of numerical interpretations of any common probability expression, with respondents splitting between near-zero and near-fifty interpretations \cite{mosteller1990quantifying, wintle2019verbal}.

More recent work examines how LLMs express and perceive uncertainty, spanning two related but distinct questions: how models interpret verbal probability language, and how reliably their expressed confidence tracks their actual accuracy or reasoning. 
Closest to our own work, Belem {\em et al.} \cite{belem2024perceptions} showed that LLMs map verbal probability expressions to numerical values in a broadly human-like manner, but that this mapping is systematically distorted by the model's prior beliefs about the associated statement. 
Specifically, they found that the same expression is assigned a higher probability when paired with a statement the model believes true than one it believes false. 
This bias is substantially larger than the corresponding effect observed in humans. 
A separate line of work examines not word-to-number mapping but the reliability of the models' expressed confidence more generally. 
Steyvers et al. \cite{steyvers2025whatllms} show that human users systematically overestimate the accuracy of LLM responses, and that longer explanations inflate users' perceived confidence independently of the model's actual accuracy. 
Steyvers and Peters \cite{steyvers2025metacognition} review this broader literature, identifying a persistent gap between models' implicit confidence, which is recoverable from token probabilities, and their explicit, verbalized confidence, with the latter being less reliable and more prone to overconfidence. 
Tanneru {\em et al.} \cite{tanneru2024quantifying} isolate one source of this gap in the context of chain-of-thought and token-importance explanations. 
The authors show that verbalized confidence scores are almost uniformly near-maximal regardless of whether the underlying answer is correct, making them uninformative, whereas an alternative probing-based measure correlates with both answer correctness and the faithfulness of the explanation. 
Our work is closest in spirit to \cite{belem2024perceptions} but differs in scope. Rather than testing whether a model's belief about a statement contaminates its report of a speaker's stated confidence, we isolate the lexical semantics of the probability words themselves, independent of speaker or statement context. This lets us compare LLM mappings directly against an established human benchmark, across a substantially larger and more current set of models, under both forced-response and explanation-elicitation conditions, and with a novel bidirectional consistency check.


\section{Experimental Setup}
\label{sec:experimental}

In this section we will describe in detail the different experimental settings we used in our study. 
We will also provide the reasoning behind these choices. 

{\bf Stimuli: } 
We use eleven verbal probability expressions spanning the full range of subjective certainty: impossible, very unlikely, improbable, unlikely, possible, even chance, probable, likely, very likely, almost certain, and certain. 
This set was chosen to follow the human benchmark established by Mosteller and Youtz \cite{mosteller1990quantifying}, allowing us to directly compare to their aggregated meta-analytic norms. 

{\bf Prompt Conditions: }
We evaluate each model under three prompting conditions. 
Each condition targets a different aspect of a model's verbal probability perception, the basic mapping itself, the effect of explicitly eliciting an explanation on that mapping, and the internal consistency of the mapping when queried bidirectionally. 

{\em (a) Basic forced-number prompt.} 
Models are given each of the eleven words in turn and asked to output a single number between 0 and 100, with the endpoints anchored: 0 corresponds to an event that will never happen, and 100 to one that always happens. 
No explanation is requested. 

\begin{tcolorbox}[colback=gray!10!white, colframe=black, title=Prompt 1, breakable]
Estimate the probability that the event described by the word \{word\} WILL HAPPEN, on a scale from 0 to 100.\newline
Scale definition:
    \begin{itemize}
        \item 0 = it will definitely NOT happen (0\% chance of occurring)
        \item 50 = it is equally likely to happen or not (a coin flip)
        \item 100 = it will definitely happen (100\% chance of occurring)
    \end{itemize}
    A HIGHER number always means MORE likely to happen. \newline
Respond with a single integer (the probability the event happens) and nothing else.
\end{tcolorbox}

This condition establishes each model's baseline word-to-number mapping under minimal instruction, analogous to the forced-response paradigm used in the human decision-science literature. 
We also make the anchoring (0 and 100) explicit and unambiguous within the instruction itself, rather than relying on the model to correctly infer scale direction. 
The reason for this, is that during some preliminary testing we observed that models sometimes invert the direction of the scale, treating higher numbers as indicating lower probability. 

{\em (b) Word-to-number mapping with explanation elicitation.} 
This prompt is the same as the previous one, with the important addition of allowing the model to explain its choice step-by-step. 
 
\begin{tcolorbox}[colback=gray!10!white, colframe=black, title=Prompt 2,breakable]
    Estimate the probability that the event described by the word \{word\} WILL HAPPEN, on a scale from 0 to 100.\newline
    Scale definition:
    \begin{itemize}
        \item 0 = it will definitely NOT happen (0\% chance of occurring)
        \item 50 = it is equally likely to happen or not (a coin flip)
        \item 100 = it will definitely happen (100\% chance of occurring)
    \end{itemize}
    A HIGHER number always means MORE likely to happen. Respond with a JSON object with exactly two fields: "number" (an integer 0-100 = the probability the event happens) and ``reason'' (a short paragraph of 3-4 sentences). In your reasoning, always phrase the probability as the chance the event DOES happen, and make sure it is consistent with "number". \newline
    Respond with valid JSON and nothing else.
\end{tcolorbox}
 
Comparing the plain and explanation-elicitation variants lets us test whether externalizing an explanation is associated with a change in the central tendency, the variance, or both, of a model's response to a given word.

{\em (c) Roundtrip (bidirectional consistency) prompt.} 
The first two conditions test whether a model can go from word to number. 
The roundtrip condition tests whether that mapping is coherent in both directions. 
We first sample a number uniformly at random from [0, 100] and ask the model to name a probability word it associates with that number. 

\begin{tcolorbox}[colback=gray!10!white, colframe=black, title=Prompt 3a, breakable]
The number \{number\} represents a probability on a scale from 0 to 100, where 0 means ``definitely will NOT happen'' and 100 means ``definitely WILL happen''. Give me a single word or short phrase (like ``likely'', ``almost certain'', ``unlikely'') that best describes this probability.\newline
Respond with only the word or phrase and nothing else.

\end{tcolorbox}

In a separate, subsequent query, we present the model with the word it just generated and ask it to assign a number to that word. 

\begin{tcolorbox}[colback=gray!10!white, colframe=black, title=Prompt 3b,breakable]
What probability (0–100) does the word or phrase \{word\} represent, where 0 means ``definitely will NOT happen'', 50 means ``equally likely either way'', and 100 means ``definitely WILL happen''? \newline
Respond with a single integer and nothing else.
\end{tcolorbox}

We define the \textbf{roundtrip error} as the absolute difference between the originally sampled number and the number recovered in
this second query. 
A model with a perfectly coherent internal
representation of probability language should recover a number close to the one it started from. 
In practice, a nonzero roundtrip error can arise from two distinct sources: a model may only ever express a small number of distinct probability values, in which case some error is unavoidable regardless of consistency, or the model's word-to-number and number-to-word mappings may genuinely disagree with each other even when each direction looks reasonable in isolation. 
Rather than treating the roundtrip error as a single measure of inconsistency, we introduce baselines in Section~\ref{sec:roundtrip} that separate these two sources.

{\bf Models: }
We evaluate the following models:

\begin{itemize}
    \item {\bf Anthropic: } Claude Haiku 4.5, Claude Sonnet 4.5, Claude Opus 4.5
    \item {\bf OpenAI: }GPT-4.1, GPT-4o, GPT-4o-mini, GPT-5, GPT-5.1, GPT-5.2, GPT-5.4, GPT-5.4-mini, GPT-5.4-nano, GPT-5.5, GPT-OSS-20B
    \item {\bf Google: }Gemma4 8B, Gemma4 26B
    \item {\bf Meta: } Llama3-8B
    \item {\bf Mistral AI: }Mistral-7B
    \item {\bf Alibaba: }Qwen-14B
\end{itemize}

This set spans both proprietary and open-weight models, ranging from lightweight variants intended for low-latency deployment (e.g., GPT-5.4-nano, Gemma4 8B, Mistral-7B) to large frontier-scale models (e.g., GPT-5, Claude Opus 4.5), allowing us to assess whether verbal probability perception is a capability that scales with model size and access tier or one that is more uniformly present across the current model landscape. 

For prompts 1 and 2 each one is sent 10 times to each model. 
For the roundtrip experiment we sample 30 numbers between 0 and 100 for each model. 
For each sampled number, we independently
repeat the full roundtrip (Prompts 3a and 3b) five times, yielding 150 roundtrip trials per model in total. 
Repeating the full roundtrip, rather than only the second step, allows both the number-to-word and word-to-number directions to vary independently across repetitions of the same input.
In all conditions, we do not set the temperature parameter explicitly, relying on each provider's API default. 
We adopt this design deliberately. 
Most users interact with these models through consumer-facing chat interfaces that do not expose or modify the temperature parameter, so the default value reflects the conditions under which verbal probability expressions are actually produced and interpreted in practice. 
We note that default temperatures are not standardized across providers and
may therefore differ across the models we evaluate.

\section{Results}
\label{sec:results}

In this section we will present the results from our experiments and discuss their implications. 

\subsection{LLMs track the human benchmark closely}

\begin{figure}[t]
    \centering
    \includegraphics[width=\linewidth]{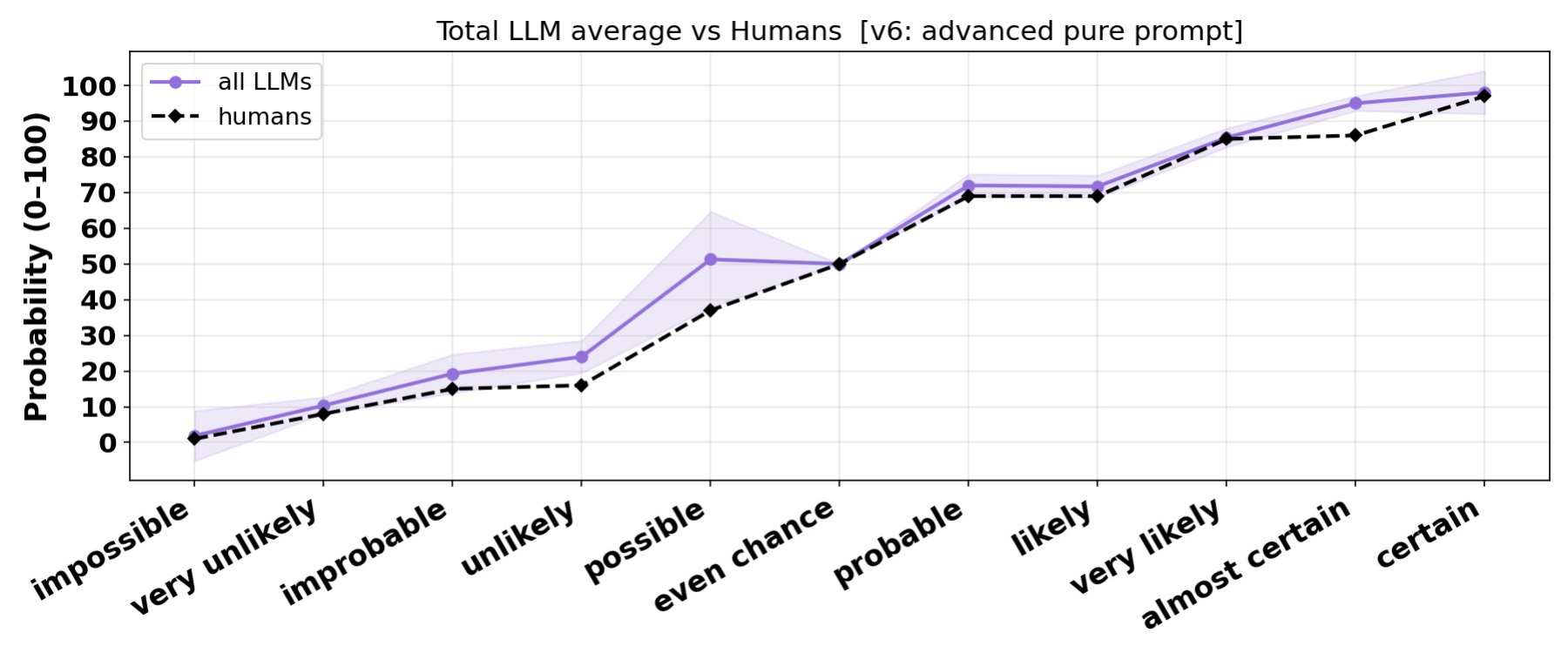}
    \caption{Aggregate probability scale across all evaluated LLMs
    versus the human benchmark, under prompt 1.
    Shaded regions denote $\pm 1$ standard deviation across models. The largest
    LLM--human divergence occurs at \textit{possible}.}
    \label{fig:llm-vs-humans}
\end{figure}

Figure~\ref{fig:llm-vs-humans} shows the aggregate mapping across all evaluated
models, using Prompt 1, against the human benchmark established by Mosteller and Youtz~\cite{mosteller1990quantifying}.
The two curves are closely aligned across most of the scale. 
In particular, the word ordering is preserved, and the three anchor expressions, namely, \textit{impossible}, \textit{even chance}, and \textit{certain}, are recovered with near-zero
variance across models, closely matching the corresponding human anchors. 
The clearest departure from the human curve occurs at \textit{possible}, where the aggregate LLM mean ($\approx$51) sits well above the human mean ($\approx$37) and carries the widest confidence band of any point on the
curve. 
Despite the mismatch, this behavior is consistent with the bimodal and unstable interpretation of this word as documented in the human decision-science literature \cite{mosteller1990quantifying,wintle2019verbal}, which we return to below. 

We also examine whether the aggregate mapping obeys approximate numeric complementarity
across antonym pairs, independent of any single word's absolute position on the scale.
For a word $w$ in our eleven-word set, let
$\mu(w)$ denote the mean numerical response assigned to $w$, averaged across all models and trials under Prompt~1 (for the LLM curve) or across all human participants (for the human benchmark curve). 
For an antonym pair $(w, \bar{w})$, e.g., (\textit{unlikely}, \textit{likely}), we define the \emph{complementarity gap} as: 
\begin{equation}
    D(w, \bar w) \;=\; \mu(w) + \mu(\bar w) - 100.
    \label{eq:complementarity}
\end{equation}
A pair with $D(w,\bar w) = 0$ is perfectly complementary, that is, the two antonyms partition the probability scale exactly in half, as would be the case if, e.g., $\mu(\text{unlikely}) = 20$ and
$\mu(\text{likely}) = 80$. 
A nonzero value of $D(w,\bar w)$ indicates that the pair's responses do not partition the probability scale into exact complements. 
This form of \emph{subadditivity}, where judgments of complementary outcomes fail to sum to the full scale, aligns with a long-documented phenomenon in the judgment-under-uncertainty literature on complementary probability estimates, most notably the finding that explicitly unpacking an event into sub-events inflates its judged probability relative to the packed description \citep{tversky1994support}. 
While the mechanism in our setting (antonym word pairs rather than event unpacking) differs, the resulting failure of complementarity we will see in what follows is qualitatively similar. 
Because $D$ is a property of the pair as a
whole, it does not indicate which member of the pair, or in which direction, departs from exact complementarity.

Of our eleven words, four pairs have a natural antonym partner: (\textit{impossible}, \textit{certain}), (\textit{very unlikely}, \textit{very likely}), (\textit{improbable}, \textit{probable}), and (\textit{unlikely}, \textit{likely}). 
The remaining three words do not have a clean antonym counterpart in our word list and are
excluded from this analysis.  
This excludes the word \textit{possible},
the word with the largest LLM--human divergence.

Table~\ref{tab:complementarity} reports
$D(w,\bar w)$ for both the aggregate LLM curve and the human benchmark.
Both populations show sub-additivity across all four pairs. 
However, LLM pairs are consistently closer to
perfect complementarity than the corresponding human pairs, most visibly for (\textit{improbable}, \textit{probable}) and
(\textit{unlikely}, \textit{likely}).

\begin{table}[t]
\centering
\begin{tabular}{lrrr}
\toprule
Antonym pair $(w, \bar w)$ & $D_{\text{LLM}}$ & $SD$ across models & $D_{\text{human}}$ \\
\midrule
impossible / certain              & $-0.07$ & 1.24 & $-2$  \\
very unlikely / very likely       & $-4.26$ & 2.89 & $-7$  \\
improbable / probable             & $-8.75$ & 6.91 & $-16$ \\
unlikely / likely                 & $-4.29$ & 5.37 & $-15$ \\
\bottomrule
\end{tabular}
\vspace{0.15in}
\caption{Antonym complementarity gap $D(w,\bar w) = \mu(w) + \mu(\bar
w) - 100$ for the 4 testable antonym pairs, using Prompt 1 across 19 models. Values closer to zero indicate greater
complementarity, while negative values indicate sub-additivity. $D_\text{LLM}$ is  pooled across trials. ($SD$: standard deviation).}
\label{tab:complementarity}
\vspace{-0.2in}
\end{table}

\subsection{Per-model variation \& negative-word bias}

\begin{figure}[t]
    \centering
    \includegraphics[width=0.9\linewidth]{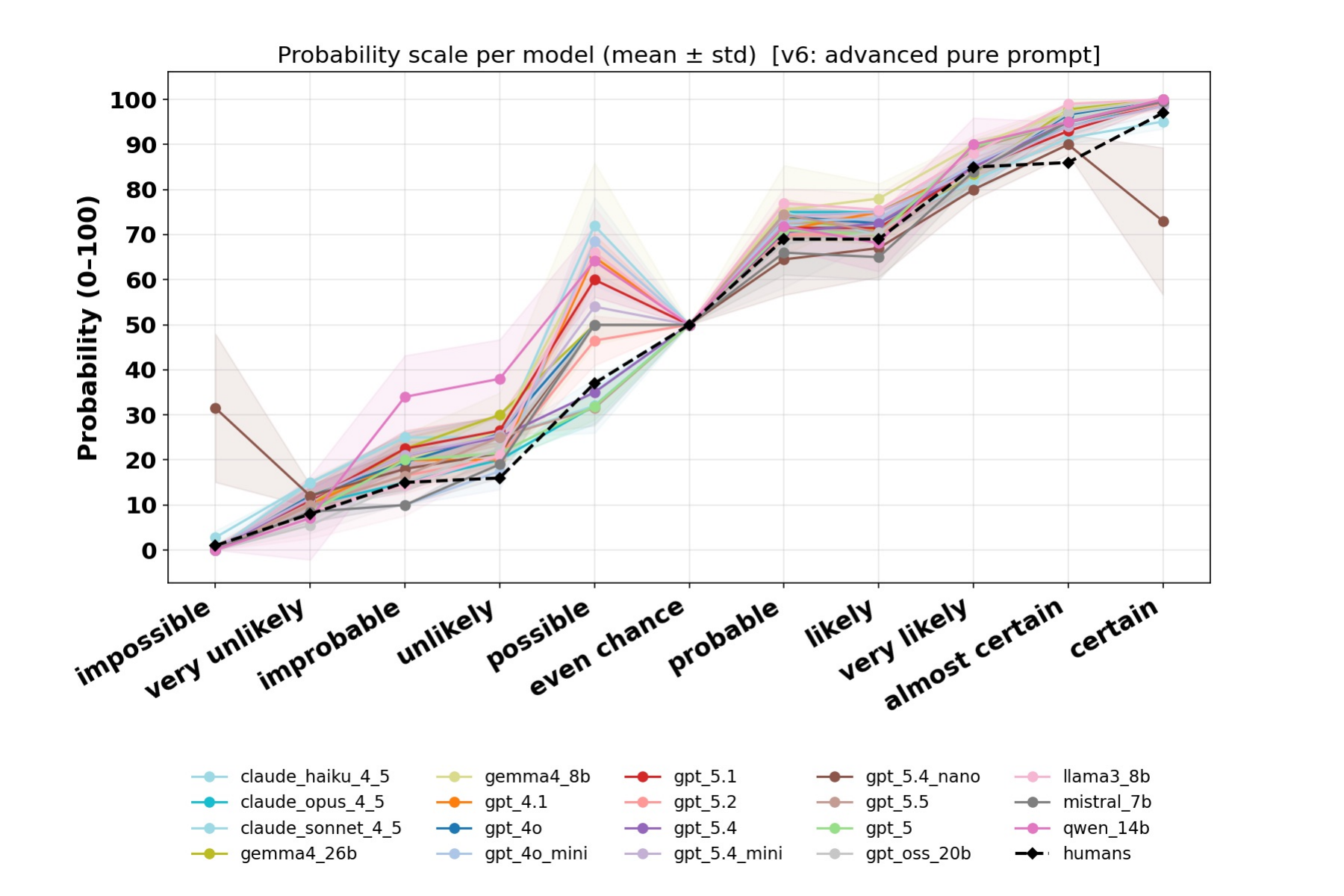}
    \caption{Probability scale per model  (Prompt 1).}
    \label{fig:line-v6}
    \vspace{-0.15in}
\end{figure}

\begin{figure}[t]
    \centering
\includegraphics[width=0.9\linewidth]{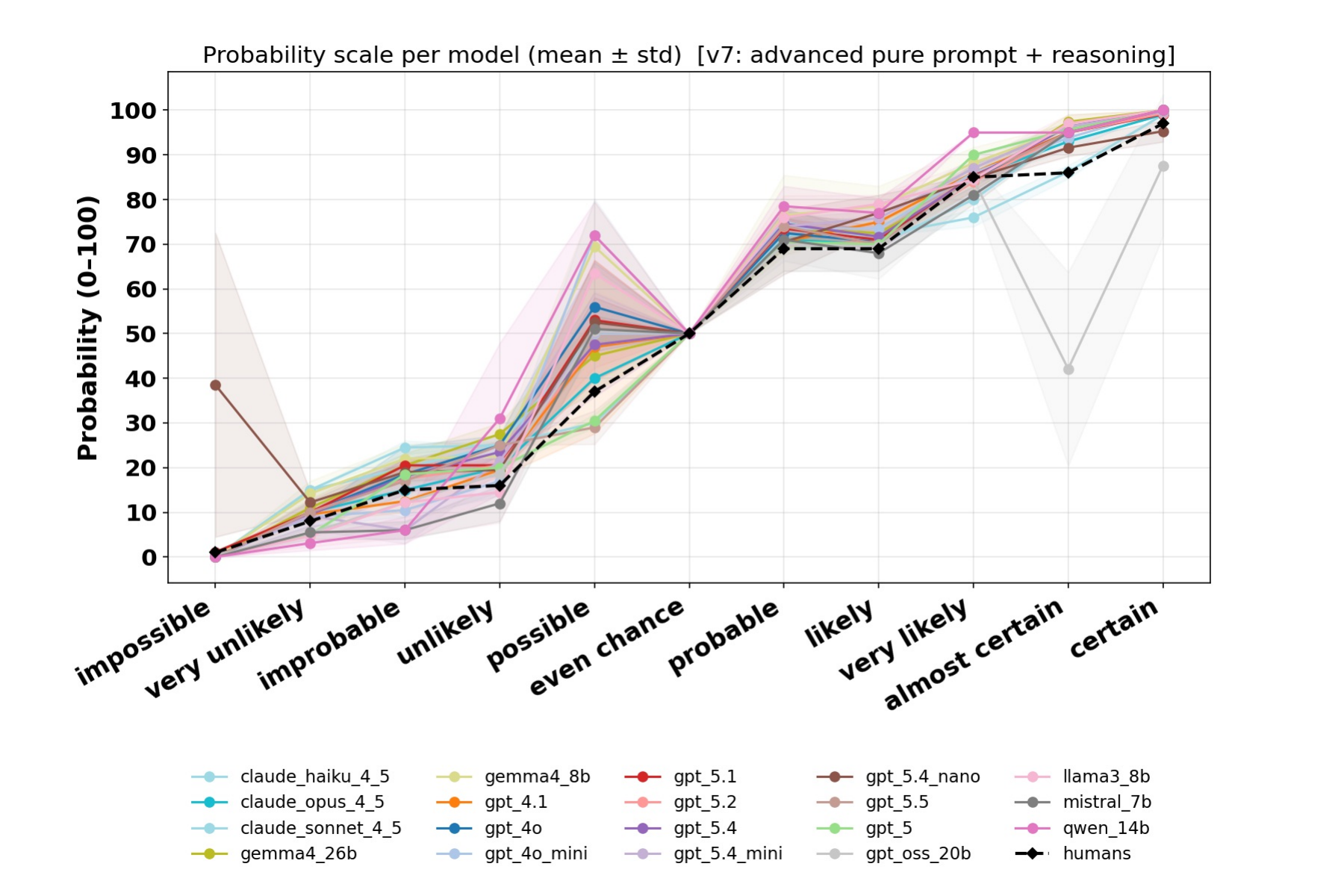}
    \caption{Probability scale per model (Prompt 2).}
    \label{fig:line-v7}
    \vspace{-0.15in}
\end{figure}

Figures~\ref{fig:line-v6} and~\ref{fig:line-v7} break the aggregate pattern down by model, comparing the LLMs' responses to Prompt 1 against the same Prompt 2, in which models are additionally asked to articulate an explanation before their final answer (explanation elicitation). 
Across both conditions, most models track the human curve closely, but do show a consistent upward shift. 
For example, \textit{unlikely} and
\textit{improbable} are assigned higher probabilities by most models than by
human respondents, with \texttt{qwen\_14b} and \texttt{llama3\_8b} showing the
largest deviations in both conditions. 
One model, \texttt{GPT-5.4 nano}, is a notable outlier at the \textit{impossible} anchor, showing substantially wider variance under explanation elicitation than under Prompt~1 (we further examine this directly
through its reasoning traces in
Section~\ref{sec:variance-exceptions}). 

Both negatively and positively worded expressions are assigned higher probabilities by LLMs than by human respondents (Wilcoxon signed-rank across 19 models, both $p<.01$ under both Prompts 1 and 2). 
However, while \textit{likely} and \textit{probable} are assigned higher values by LLMs than by humans, the gap margin is smaller than that of their negative counterparts (\textit{unlikely}, \textit{improbable}). 
Testing this asymmetry directly we find that under Prompt 1 the difference is on average about 2.1 percentage points over all 19 models (p-value for paired t-test 0.05), with a median of 0.9 (Wilcoxon signed-rank test $p=.067$). 
Under Prompt 2 the differences are not significant. 
Given the sample size for these tests we can clearly see that they are underpowered to detect small to moderate effects.   
We therefore treat this asymmetry itself as suggestive rather than established. 
This is consistent with the antonym-complementarity result we saw earlier in Section~\ref{sec:results} (Table~\ref{tab:complementarity}). 
Both members of a pair shift upward by similar amounts, keeping LLM pairs closer to perfect complementarity than human pairs. 
However, as we can see from the standard deviation (SD) column in Table~\ref{tab:complementarity}, there is substantial heterogeneity across individual models. 

Comparing the results from the two prompts directly, we see that explanation elicitation visibly tightens the variance bands for most models, 
but it does not eliminate the upward shift. 
Furthermore, for a subset of models, mainly smaller open-weight models such as \texttt{qwen\_14b} and \texttt{llama3\_8b}, explanation elicitation appears to \emph{increase} divergence from the human curve at \textit{possible} and \textit{probable} rather than reduce it. 
This suggests that explanation elicitation reduces the variance of each model's \emph{individual} response distribution without necessarily pulling models toward consensus with either humans or each other. 
This instability effect seems to also be more pronounced in smaller, open-weight models than in frontier proprietary ones.
While the polarity asymmetry we examined earlier is not robustly established, a paired comparison reveals a distinct and better-powered result. 
Each model's own gap is correlated across the two prompts ($r=.64$ across 19 models), so
comparing a model against \emph{itself} rather than against the population average removes much of the between-model noise that limits the two marginal tests. 
Under this paired comparison, the negative/positive gap shrinks under
explanation elicitation for 13 of 19 models (Wilcoxon signed-rank $p=.006$). 
This indicates that whatever polarity asymmetry a given model exhibits, explanation elicitation tends to reduce it. 
This is a within-model claim that is statistically distinct from, and
better supported than, the across-model, population-level question of
whether the asymmetry exists at all.


\subsection{Explanation elicitation does not uniformly reduce variance}
\label{sec:variance-exceptions}

\begin{figure*}[t]
    \centering
    \begin{subfigure}[b]{0.29\linewidth}
        \centering
        \includegraphics[width=\linewidth]{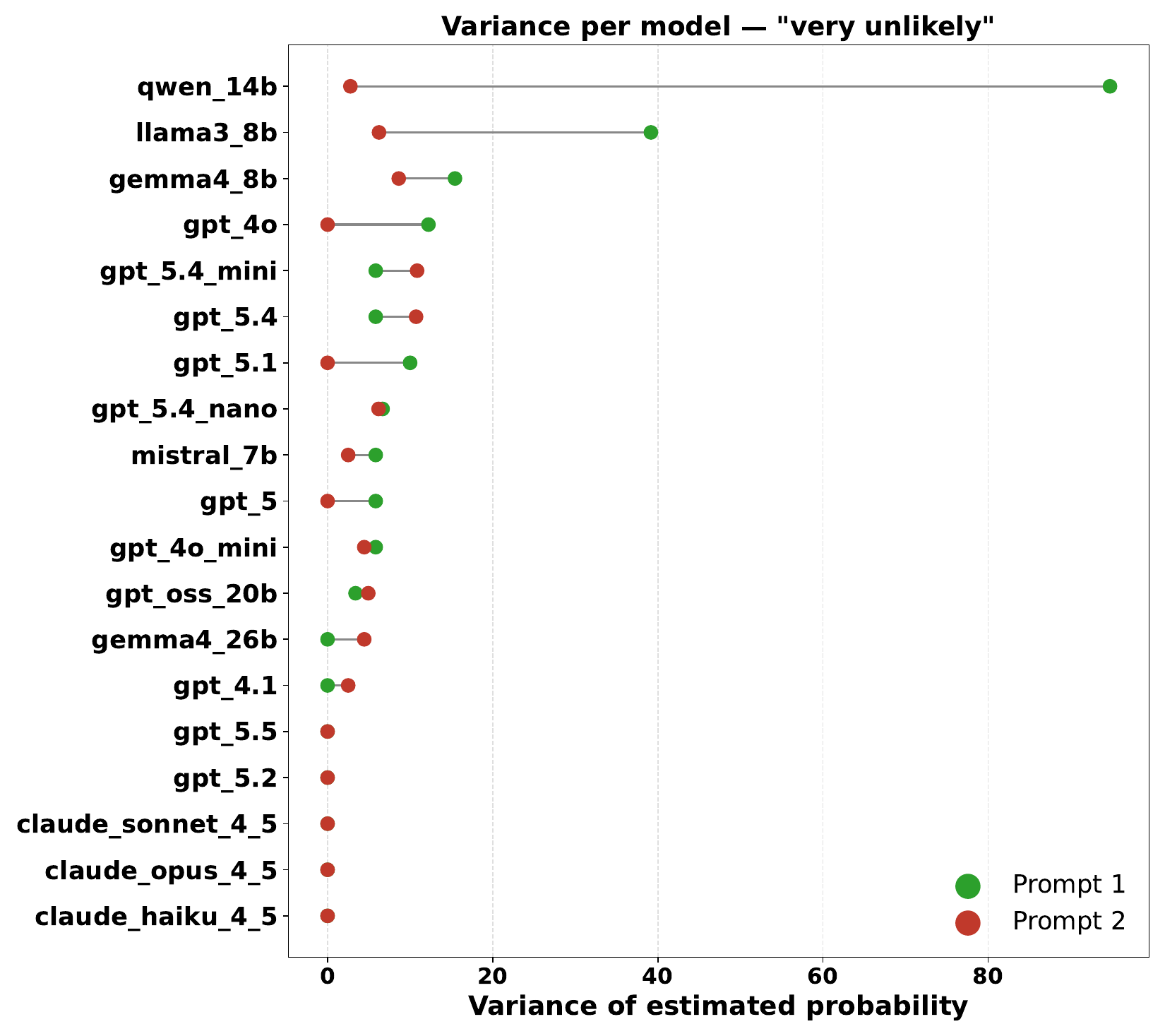}
        \caption{\textit{very unlikely}}
        \label{fig:var-very-unlikely}
    \end{subfigure}
    \hfill
    \begin{subfigure}[b]{0.29\linewidth}
        \centering
        \includegraphics[width=\linewidth]{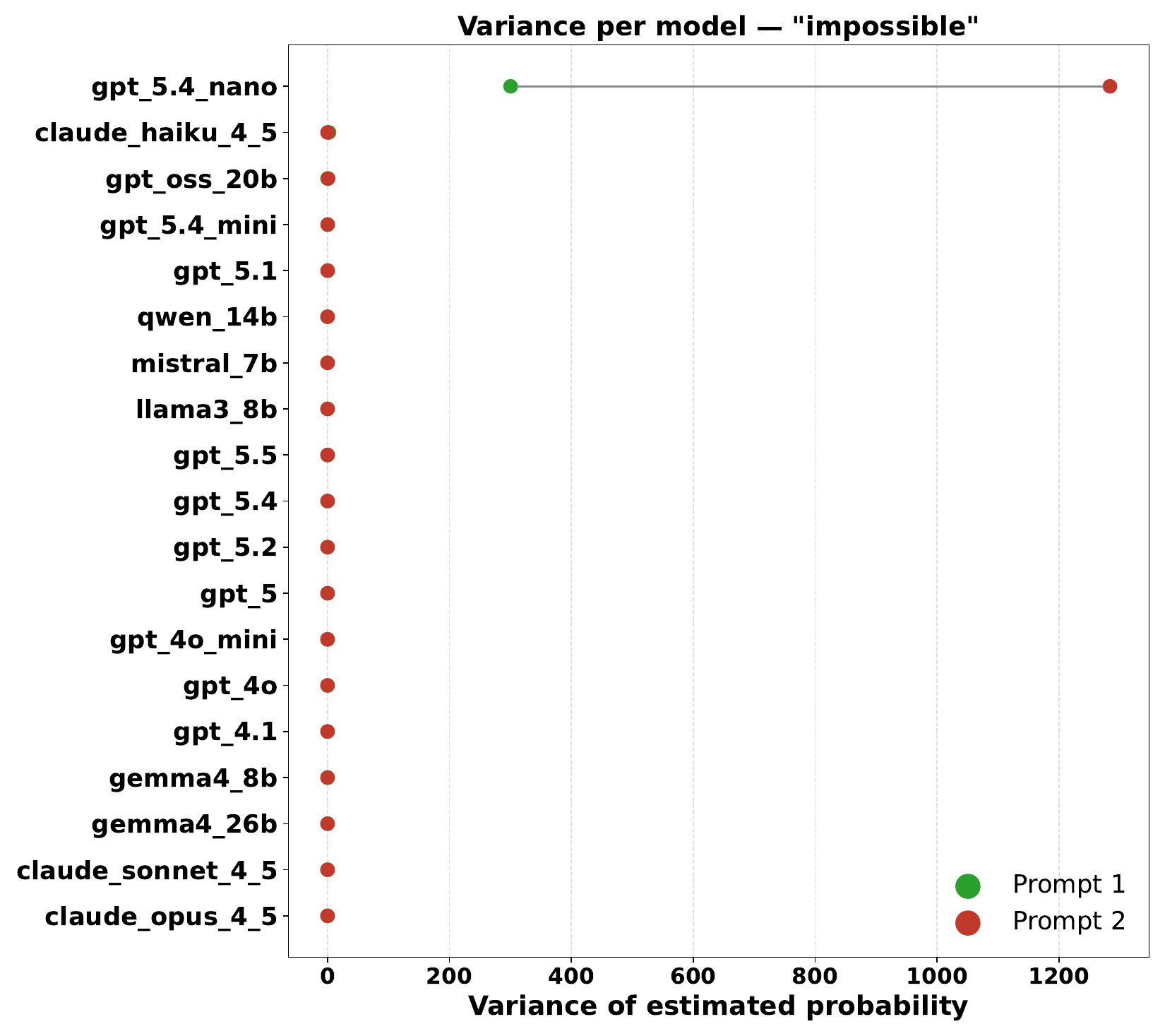}
        \caption{\textit{impossible}}
        \label{fig:var-impossible}
    \end{subfigure}
    \hfill
    \begin{subfigure}[b]{0.29\linewidth}
        \centering
        \includegraphics[width=\linewidth]{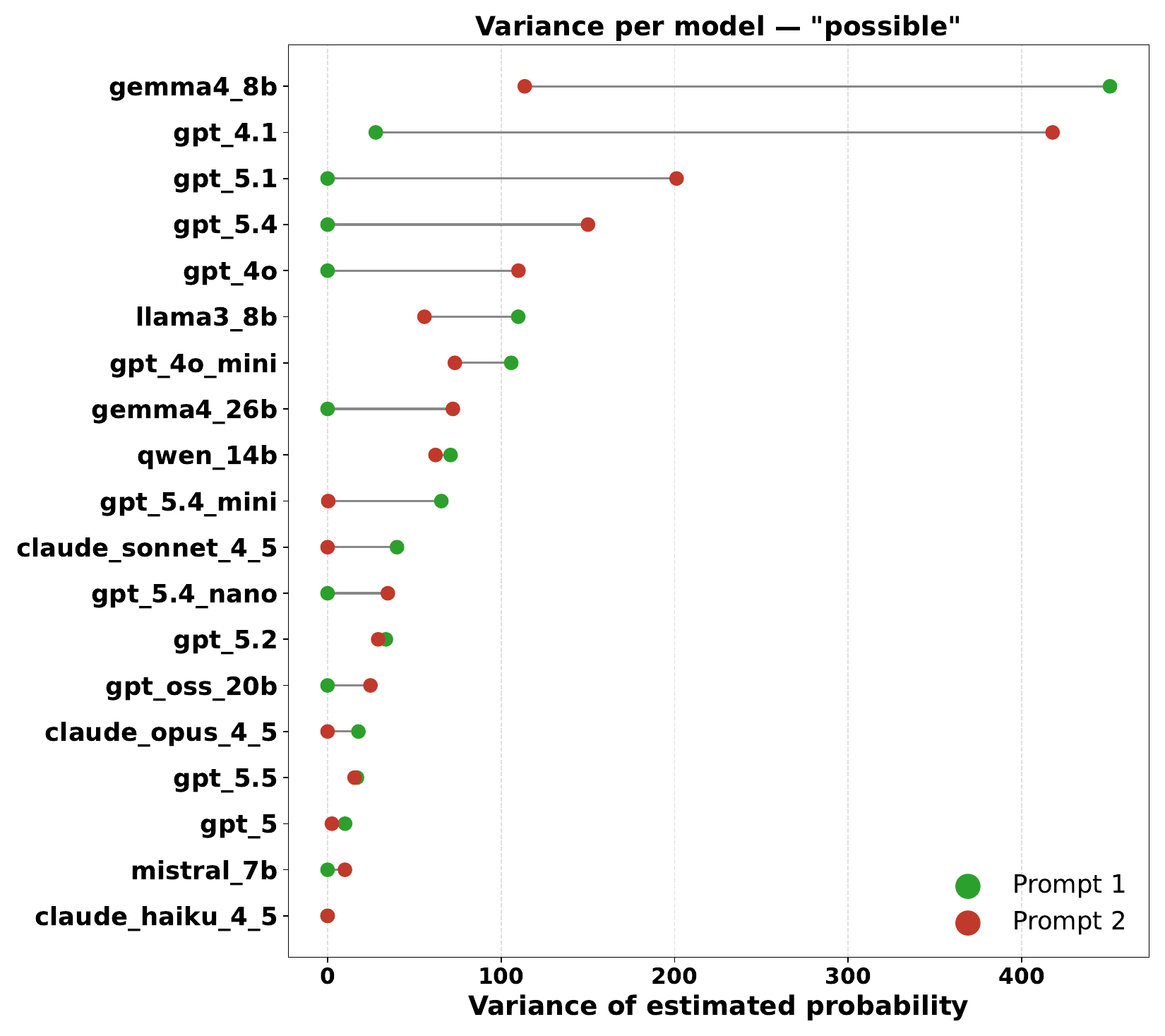}
        \caption{\textit{possible}}
        \label{fig:var-possible}
    \end{subfigure}
    \vspace{-0.1in}
    \caption{Variance of estimated probability per model, comparing
    Prompt~1 (no explanation, green) and Prompt~2 (explanation
    elicitation, red).
    (a)~\textit{very unlikely} shows the general stabilization pattern.
    Every model with substantial baseline variance shows a clear
    reduction under explanation elicitation. (b)~\texttt{gpt\_5.4\_nano} shows a
    sharp variance increase at \textit{impossible}, driven by
    within-trial scale instability. (c)~Several frontier models show
    increased variance at \textit{possible} despite near-zero variance
    elsewhere, consistent with explanation elicitation surfacing latent
    disagreement documented for this word in humans.}
    \label{fig:var-grid}
\end{figure*}

Explanation elicitation (Prompt~2) tends to reduce response variance relative to Prompt~1.
This effect is clearest for words such as \textit{very unlikely} (Figure~\ref{fig:var-very-unlikely}), where every model with
substantial baseline variance under Prompt~1 
shows a clear reduction in variance under Prompt~2.
The few cases in which variance appears to increase occur only among models whose variance is already negligible under both conditions.
However there are exceptions, which are informative. 
There are two distinct patterns in which explanation elicitation is associated with \emph{increased} variance, examined through qualitative inspection of the underlying reasoning traces.

\paragraph{Within-trial scale instability.} For \texttt{GPT-5.4-nano} at \textit{impossible}, variance increases substantially from Prompt~1
to Prompt~2 (Figure~\ref{fig:var-impossible}).
Reading the individual traces shows the model arriving at incompatible interpretations of the same instruction across trials.
Some responses correctly reason toward
a near-zero probability, others default to {\tt``50\%, given no additional
context''} and at least one trial explicitly inverts the anchoring, reasoning that {\tt ``the prompt explicitly sets up the scenario to
evaluate [$\ldots$] I estimate a 90\% chance''. }
This is the scale-direction confusion our anchored prompt design is intended to prevent.
However, here it re-emerges specifically \emph{when the model is asked to explain itself}. 
This confusion appears inconsistently across trials rather than as a fixed per-model error.
A related but distinct failure mode
occurs for \texttt{gpt-oss-20b} at
\textit{certain} and \textit{almost certain}.
The model loses track of the target task itself when asked to explain its answer.
In particular, in several trials this model evaluates the probability that the \emph{prompt text itself} is naturally-occurring language, rather than the probability denoted by the target word. 
For instance, one trial's full reasoning reads:
\begin{quote}
\ttfamily\small
``The phrase involves a nested quotation structure that would almost
never appear in ordinary text unless typed deliberately or by
mistake. Such an exact pattern is highly unlikely to occur naturally,
making the event very rare. Consequently, the probability of this
specific word sequence happening is only about one percent. Thus, it
is considered quite improbable but not impossible.''
\end{quote}
Here, the model assigns a probability of $1$, not because it judges \textit{almost certain} to denote a low probability, but because it is evaluating how likely the \emph{literal prompt text it received}, including its quotation marks, is to occur as a naturally-typed string.
The target word itself never enters the
model's explanation.

\paragraph{Latent disagreement surfaced by explanation elicitation.} For several frontier models (e.g., \texttt{gpt\_5.1}, \texttt{gpt\_5.4} etc.) variance at \textit{possible} increases sharply under explanation elicitation, even though these same models show near-zero variance at nearly every other word regardless of condition (Figure~\ref{fig:var-possible}). 
Unlike the previous pattern, this points to a principled rather than an arbitrary situation. 
The word \textit{possible} is documented in the human decision-science literature as eliciting an unusually unstable, bimodal split in
interpretation \citep{mosteller1990quantifying,wintle2019verbal}. 
Asking these models to explain their answer may lead them to surface and act on multiple candidate interpretations across different
trials, rather than collapsing to a single default response as they do under Prompt~1. 
We note that this is consistent with, but does not by itself establish, a directly parallel bimodal tendency in these
models. 
Our data show increased \emph{variance} across trials at this
word, but we have not formally tested for multimodality (e.g., via a
mixture-model fit) given our sample size per model.

Together, these patterns indicate that the variance-reducing effect of explanation elicitation is not uniform.
Specifically, it depends on the word
being evaluated, and on whether articulating an explanation supports genuine resolution of ambiguity (as with \textit{possible}) or instead introduces new opportunities for the model to misconstrue the task or invert the intended scale.

\subsection{Roundtrip consistency}
\label{sec:roundtrip}

We next performed the roundtrip experiment described in Section~\ref{sec:experimental} (Prompts 3a and 3b).
For each model, we sample 30 numbers uniformly from $[0,100]$ and repeat the full roundtrip five times for each sampled number, yielding 150 roundtrip trials per model in total. 
We refer to the five repeated trials sharing a given sampled input as a \emph{block}.  
Each model ends up with 26 blocks each, instead of 30 blocks, since we sample integers in this range and our sampling yields 26 distinct sampled values. 

For each model $m$, let $x_i \sim \mathrm{Uniform}[0,100]$ be the
$i$-th sampled probability presented in the first roundtrip query (3a), let $w_i = f_m(x_i)$ denote the word returned by model $m$ for that value, and let $\hat{x}_i = g_m(w_i)$ denote the number model $m$ assigns to that same word $w_i$ when queried independently in the second roundtrip step (3b). 
We define the roundtrip error for trial
$i$ as:
\begin{equation}
    e_i = \lvert x_i - \hat{x}_i \rvert,
    \label{eq:roundtrip-error}
\end{equation}
and the \emph{mean absolute roundtrip error} for model $m$ as:
\begin{equation}
    \mathrm{MARE}_m = \frac{1}{N} \sum_{i=1}^{N} e_i,
    \label{eq:mare}
\end{equation}
where $N=150$. 
A model with a perfectly self-consistent word--number mapping would achieve $\mathrm{MARE}_m = 0$.

\paragraph{Baselines.} A natural first reference point is a naive independence baseline, obtained by treating both $x_i$ and
$\hat{x}_i$ as independent draws from $\mathrm{Uniform}[0,100]$. 
This yields an expected error that is equal to the expected absolute difference of two independent uniform random variables, that is, $100/3 \approx 33.3$. 
However, this baseline is not meaningful in practice. 
It implicitly assumes the recovered value $\hat{x}_i$ is itself uniformly distributed, which no model in our data satisfies. 
A model that simply ignores the input and always answers ``equally likely'' (recovering 50 every time) achieves $\mathbb{E}|X-50|=25$ for $X\sim\mathrm{Uniform}[0,100]$, thus, beating the naive baseline of 33.3 while carrying absolutely no information about the input. 
We therefore report two corrected reference points instead: 
(a) the \emph{best-constant baseline} of 25, and, (b) a \emph{quantization floor} of $25/k$ for a model that only ever
produces $k$ distinct recovered values, representing the unavoidable error from having a limited vocabulary even if that vocabulary is used with perfect internal consistency (the derivation of the quantization floor is in
Appendix~\ref{app:quantization-floor}). 
Any excess of a model's observed $\mathrm{MARE}_m$ above its own quantization floor reflects genuine inconsistency in \emph{how} it uses its available vocabulary,
not the size of that vocabulary itself. 
We also report a permutation-based test of whether each model's observed MARE is distinguishable from chance. 
In particular, we shuffle the pairing between each model's own sampled inputs and
recovered values 5,000 times, preserving the model's actual empirical output distribution, and report the fraction of shuffles that achieve a MARE at least as low as the model's observed value. 
A large fraction here indicates that a model's roundtrip mapping is statistically indistinguishable from chance, given only its own tendency to produce certain output values regardless of input.

\paragraph{Results.} Table~\ref{tab:roundtrip} reports, for each model, its empirical vocabulary size $k$ (the number of distinct recovered values observed across all 150 trials), the resulting quantization floor, the observed MARE, the gap between the two, and
the permutation test result. 

\begin{table}[t]
\centering
\begin{tabular}{lrrrrr}
\toprule
Model & $k$ & Floor & MARE & Gap & $p$ \\
\midrule
gpt\_5.5           & 15 & 1.67 & 4.49  & 2.82  & $<$.0001 \\
gpt\_5             & 17 & 1.47 & 4.63  & 3.16  & $<$.0001 \\
gpt\_5.4           & 9  & 2.78 & 5.02  & 2.24  & $<$.0001 \\
gpt\_5.1           & 19 & 1.32 & 5.19  & 3.87  & $<$.0001 \\
claude\_sonnet\_4\_5 & 9  & 2.78 & 6.02  & 3.24  & $<$.0001 \\
claude\_opus\_4\_5  & 15 & 1.67 & 6.63  & 4.97  & $<$.0001 \\
gpt\_4.1           & 14 & 1.79 & 6.74  & 4.95  & $<$.0001 \\
gpt\_4o            & 14 & 1.79 & 7.05  & 5.26  & $<$.0001 \\
gpt\_5.2           & 13 & 1.92 & 7.15  & 5.23  & $<$.0001 \\
gemma4\_26b        & 17 & 1.47 & 7.61  & 6.14  & $<$.0001 \\
gpt\_oss\_20b      & 23 & 1.09 & 8.27  & 7.18  & $<$.0001 \\
gpt\_5.4\_mini     & 15 & 1.67 & 8.98  & 7.31  & $<$.0001 \\
gpt\_4o\_mini      & 12 & 2.08 & 12.81 & 10.72 & $<$.0001 \\
claude\_haiku\_4\_5 & 14 & 1.79 & 15.02 & 13.23 & $<$.0001 \\
gemma4\_8b         & 20 & 1.25 & 20.69 & 19.44 & $<$.0001 \\
qwen\_14b          & 8  & 3.12 & 26.31 & 23.19 & .0001 \\
gpt\_5.4\_nano     & 10 & 2.50 & 26.77 & 24.27 & $<$.0001 \\
mistral\_7b        & 8  & 3.12 & 27.01 & 23.89 & \textbf{.1024} \\
llama3\_8b         & 12 & 2.08 & 27.71 & 25.63 & \textbf{.0090}$^\dagger$ \\
\bottomrule
\end{tabular}
\vspace{0.15in}
\caption{Roundtrip consistency results across all 19 models, ranked by observed MARE. $p$ is the permutation-test $p$-value (5K
resamples). Bold values do not survive Bonferroni correction across
the 19 models tested; $\dagger$ indicates a value that survives the less conservative Holm-Bonferroni correction.}
\label{tab:roundtrip}
\end{table}

Two patterns are visible that a raw MARE ranking alone would obscure. 
First, only four models, namely,  \texttt{GPT-5.4 nano}, \texttt{Qwen-14b}, \texttt{Mistral-7B}, and \texttt{llama3-8b}, fail to beat the constant baseline of 25.  
Also, \texttt{Gemma4-8B}, despite the largest gap in the table, still achieves a lower MARE than a strategy that ignores the input entirely. 
Second, the permutation test separates these four worst-performing models from one another despite their similar raw MARE values. 
Given that we run this test across 19 models, we apply a Bonferroni correction ($\alpha/19\approx.0026$) to guard
against false positives from multiple testing. 
Under this correction, \texttt{Qwen-14b} and \texttt{GPT-5.4 nano} remain clearly distinguishable from chance. 
Essentially, whatever these two models are doing is not fully random, even though they perform worse than ignoring the input. 
Connecting to our earlier results,  \texttt{GPT-5.4 nano} is the same model whose reasoning traces we examined in Section~\ref{sec:variance-exceptions} for scale-direction instability at \textit{impossible}. 
Its roundtrip failure here is consistent with
that finding, though driven specifically by its word-to-number mapping (Step 3b) rather than its choice of word (Step 3a), where as we discuss in what follows its distinct-word ratio of .30 is only mid-pack (see Table~\ref{tab:word-stability}).
\texttt{llama3-8b} performance is borderline, as it survives a sequential Holm-Bonferroni correction ($p=.009$) but not the stricter flat Bonferroni threshold, so we treat its apparent coupling as suggestive rather than conclusive.
Finally, \texttt{Mistral-7B}'s mapping is not distinguishable from chance under any correction ($p=.102$). 
That is, we cannot
statistically distinguish this model's roundtrip mapping from one
that ignores the input entirely.

We further measure how many distinct words a model returns in Step 3a across the five trials within a block, normalized by the number of trials in that block to account for the repeated sampling of 4 out of 30 initial integers.
This diagnostic targets a different property than the empirical vocabulary size $k$ used in our quantization-floor analysis. 
Whereas $k$ counts distinct \emph{final} recovered numbers pooled across all
trials, the word-choice diversity ratio is computed \emph{within} a block of trials sharing an identical input, so any variation it detects reflects pure inconsistency rather than legitimate resolution across distinct inputs.
As we can see in Table~\ref{tab:word-stability}, \texttt{GPT-5.4}, \texttt{GPT-4.1} and \texttt{Claude-Opus-4.5} pick nearly the same word every time (a mean distinct-word ratio of .20, .21 and .22, respectively), while \texttt{Gemma4-8B} shows the least stable word choice of any model in our results (.59). 
Despite this, \texttt{Gemma4-8B} still beats the constant baseline overall, indicating that its number-mapping step (3b) compensates for highly unstable word choice.

\begin{table}[ht]
\centering
\begin{tabular}{lr}
\toprule
Model & Mean distinct-word ratio \\
\midrule
gpt\_5.4            & .196 \\
gpt\_4.1            & .208 \\
claude\_opus\_4\_5   & .219 \\
gpt\_5.2            & .242 \\
gpt\_4o\_mini        & .246 \\
gemma4\_26b         & .250 \\
gpt\_4o             & .250 \\
claude\_sonnet\_4\_5 & .258 \\
gpt\_5.1            & .273 \\
gpt\_5.4\_mini       & .277 \\
gpt\_5.5            & .277 \\
claude\_haiku\_4\_5  & .285 \\
gpt\_5              & .285 \\
gpt\_5.4\_nano       & .300 \\
qwen\_14b           & .350 \\
mistral\_7b         & .354 \\
gpt\_oss\_20b       & .427 \\
llama3\_8b          & .438 \\
gemma4\_8b          & .588 \\
\bottomrule
\end{tabular}
\vspace{0.15in}
\caption{Mean distinct-word ratio per block (five trials sharing an identical sampled input, normalized by block size), ranked from most to least stable, across all 19 models.}
\label{tab:word-stability}
\vspace{-0.15in}
\end{table}

\paragraph{Within-family comparisons.} Focusing on models from the same family allows us to isolate the effect of scale from confounds of
architecture and training pipeline. 
Of course, this isolation is not perfect, since same-family variants can still differ in fine-tuning data and other choices beyond parameter count, but nonetheless is informative. 
Within the Gemma4 family, the 26B variant achieves a substantially lower roundtrip error than the 8B variant (7.61 vs.\ 20.69). 
A similar, but less pronounced, pattern holds
within the GPT-4o family, where the full-size model outperforms its mini variant (7.05 vs.\ 12.81). 
Within the GPT-5 family, we restrict this comparison to the three variants whose naming denotes an explicit size tier (\texttt{gpt\_5}, \texttt{gpt\_5.4\_mini}, \texttt{gpt\_5.4\_nano}). 
Roundtrip error increases monotonically
across the three size tiers: 4.63, 8.98, 26.77. 
Even the same-tier point releases (\texttt{gpt\_5.1}, \texttt{gpt\_5.2}, \texttt{gpt\_5.4}, \texttt{gpt\_5.5}), all  cluster tightly between 4.49 and 7.15 regardless of version number. 
These are all lower than  \texttt{gpt\_5.4\_mini} and  \texttt{gpt\_5.4\_nano}, and consistent with these point releases being updates within the same size tier rather than different model sizes. 
The three Claude variants follow a similar, though not perfectly monotonic, ordering by size (\texttt{claude\_opus\_4\_5}: 6.63, \texttt{claude\_sonnet\_4\_5}: 6.02, \texttt{claude\_haiku\_4\_5}: 15.02).

These comparisons provide more direct evidence for an effect of scale on roundtrip consistency than cross-family comparisons alone, where
architecture and training differences could otherwise account for the observed gap. 
We note, however, that with only a few models per family, we cannot distinguish a robust scale effect from incidental variation, and a larger sample of same-family model sizes would be needed to establish this more rigorously.

\section{Discussion and Limitations}
\label{sec:discussion} 

Taken together, our results point to a consistent picture, where LLMs inherit the ``structure'' of human verbal probability perception without fully inheriting its calibration. 
More specifically, word ordering and anchor fidelity are reproduced across nearly every model we evaluate, regardless of scale or
provider, and the highest variance and cross-model disagreement in our results occur specifically at \textit{possible}, which is consistent with the bimodal instability documented for this word in humans.
This suggests that this structure is a robust and learned property of the distribution of probability language in pretraining data rather than a capability that requires scale or explicit alignment. 
At the same time, models diverge from humans in a specific and repeatable way. 
LLMs systematically rate both negatively and
positively worded expressions as more probable than human norms suggest, with a marginal trend toward this inflation being larger for negative expressions specifically. 
While explanation elicitation improves each model's \emph{internal} consistency (narrower
response variance), it does not correct this bias and, for a subset of models, appears to increase divergence from the human curve rather
than reduce it. 
This might serve as an indication that eliciting explanation stabilizes a model's individual mapping without necessarily correcting it or moving models toward consensus with one another. 
This connects to a broader concern raised in prior work on LLM reasoning outside the probability domain. 
In particular, chain-of-thought prompting has been found to be able to induce behavioral shifts, such as collapsing onto a narrow set of answer patterns, that are difficult to distinguish from genuine improvements in task understanding \citep{zevcevic2023causal}. 
Our transcript-level evidence in Section~\ref{sec:variance-exceptions} suggests a similar risk applies to reasoning about verbal probability expressions. 
For at least two models, the instructions to provide an explanation appear to introduce new failure modes (task misidentification and within-trial scale inversion) rather than resolving genuine uncertainty about the
target word. 

The roundtrip experiment aims at quantifying the self-consistency of each model, i.e., whether a model's word-to-number and number-to-word mappings agree with each other, rather than at tracking calibration accuracy (i.e., how closely those mappings match human values). 
Most models maintain a coherent bidirectional representation of probability language, while a small subset, spanning both proprietary and open-weight families, show mappings that are
statistically indistinguishable from chance. 
Within-family comparisons (e.g., Gemma4 26B vs.\ 8B, GPT-4o vs.\ GPT-4o-mini, and a
three-tier ladder within the GPT-5 family) suggest that this self-consistency gap is at least partly attributable to scale, independent of architecture or training pipeline.

{\bf Human-AI communication: }The positive bias documented above, and specifically the potentially larger inflation for negatively worded expressions, has a direct practical consequence.
This bias operates on the grammatical polarity of the expression itself, independent of whether the event being described is desirable or undesirable.  
A human reader will tend to 
\emph{underestimate} the probability a model actually assigns whenever it hedges using negative wording, regardless of whether that wording describes an undesirable event (``side effects are \textit{unlikely}''), 
in which case the reader is given false reassurance, or a desirable one (``recovery is \textit{unlikely}''), in which case the reader is led to be more pessimistic than the model's own estimate warrants. 
Positive-worded hedges (\textit{likely}, \textit{probable}) show the same directional bias but to a smaller degree, so the risk of underestimation is reduced, not eliminated, when a model hedges using positive language.
In domains where such hedges carry real consequences, this systematic inflation is very important.  
Since a model's own mapping for a hedge word sits above the corresponding human mapping, this can lead to the probability underestimation mentioned above. 
Whether this is false reassurance or undersold good news depends on the desirability of
the event being described. 
A clinical assistant describing side effects as \textit{unlikely}, or a safety system reporting a fault as \textit{improbable}, risks leaving a
patient or operator more confident in a good outcome than the model's own estimate warrants. 
In the opposite direction, a financial tool describing a recovery as \textit{likely}, or a forecasting system describing an opportunity as \textit{probable}, risks understating the model's own confidence in a desirable outcome, leading a user to discount good news more than the model's estimate would justify. 
This is a distinct mechanism from the reliability-overestimation effects documented elsewhere in the literature, where users trust an LLM's correctness more than its actual accuracy warrants, often driven by superficial cues such as response length
\citep{steyvers2025whatllms}. 
The bias we report does not concern whether a model's underlying claim is accurate, but whether the numeric belief a model intends to convey through a hedge word is the same as that of the human user. 
Even a model with perfectly calibrated internal probabilities could, through this word-choice mismatch alone, leave a human reader
with a systematically different estimate than the model's own internal state would justify. 

{\bf Limitations: }As aforementioned, we query each model at its default temperature setting  in order to reflect how most users interact with these systems through consumer-facing interfaces. 
However, default temperatures are not standardized across providers, so some of the
cross-model variance we report may be partly confounded with each model's baseline stochasticity rather than reflecting a pure effect of reasoning or model tier. 
Additionally, each triplet (model, word, condition) is estimated from only $N=10$
samples. 
While we believe that this is sufficient to establish the directional patterns we
report, individual per-word means, especially for the smaller open-weight models with wider variance, should be interpreted with
appropriate uncertainty. 
Finally, our forced-response and explanation elicitation prompts use a single fixed instruction template. 
We have not tested robustness to paraphrasing the anchoring instruction itself, so we cannot rule out that some fraction of the observed bias is sensitive to the specific wording we chose rather than being a fully prompt-invariant property of these models.

Our human benchmark is drawn from the meta-analytic norms of Mosteller and Youtz~\cite{mosteller1990quantifying}, aggregated across studies conducted well before the emergence of current LLMs. 
Since we rely on this existing benchmark,
some of the divergence we attribute to LLMs could in principle reflect true shifts in population-level human interpretation since these norms were established, rather than a property specific to LLMs. 
Finally, our findings are necessarily specific to the eleven expressions, three prompt conditions, and the models we evaluate. 
Extending this analysis to a broader vocabulary of uncertainty language, additional prompt phrasings, and the continually evolving landscape of available models is an important direction for future work. 
A further extension worth pursuing is \emph{comparative} and \emph{contextual} probability language, which convey a probability \emph{relative}
to some reference point rather than an absolute value. 
This includes expressions such as
\textit{more likely than}, \textit{slightly more probable}, or
\textit{far less likely}.  
The same comparative phrase can correspond to very different absolute magnitudes depending on context (e.g., ``more likely'' could indicate a shift from 10\% to 15\%, or from 40\% to 90\%). 
This is a form of ambiguity entirely distinct from the context-free lexical mapping we study here, and one that may interact with the biases we report in ways that context-free elicitation cannot reveal. 

Finally, two further metrics would strengthen the results but require substantially more data than we collected in this study. 
First, an entropy-based measure of word-choice consistency would be more informative than the distinct-word ratio, since it could distinguish a model that splits its answers at a ratio of 4-to-1 between two words from one that splits them evenly. 
Second, a mutual information measure between the sampled input and the word or number returned would directly quantify how much information survives the roundtrip, incorporating much of what $k$, the quantization floor and the permutation test currently do separately into a single
statistic. 
Both metrics would be feasible with
data on the order of thousands, rather than hundreds, of roundtrip trials per model, which we identify as a concrete direction for future work.

\section{Conclusions}
\label{sec:conclusions}

LLMs largely inherit the structure of human verbal probability perception (word ordering, anchor fidelity, a rich vocabulary in frontier models) but not its calibration. 
Both negatively and positively worded expressions are overrated relative to human norms, with only a marginal, non-robust trend toward this being larger for negative expressions specifically. 
Also, neither explanation elicitation
nor scale reliably fixes the underlying inflation. 
Our experiments also indicate that self-consistency and calibration accuracy are
separate, only loosely related properties.
These results suggest that current LLMs communicate uncertainty in a recognizably human-like way on the surface, but carry specific, measurable distortions that matter wherever hedged language is trusted at face value. 

\small
\bibliographystyle{ACM-Reference-Format}
\bibliography{sample-base}  

\appendix

\section{Quantization Floor Derivation}
\label{app:quantization-floor}

During the roundtrip experiment we noted that some models recover only a small number of distinct values across the roundtrip  procedure, regardless of the original sampled input. 
For example, a model might only ever express its answer using a handful of familiar
round-number anchors (e.g., 25, 50, 75, and 95, say) rather than the full continuum between 0 and 100. 
Such a model cannot achieve zero roundtrip error even if it is otherwise perfectly consistent. 
For example, if the original input was 43, and the model always rounds to the nearest of its four available anchors (50), it will report 50 regardless of how many times the roundtrip is repeated. 
This is not evidence of an incoherent mapping, but rather it is an unavoidable consequence of having only a few values {\em available} to report. 
We derive here the smallest error such a restriction can produce, so that a model's excess error above this bound can be attributed to genuine inconsistency rather than vocabulary size.

Formally, a model that only ever reports one of $k$ distinct values is equivalent, for this purpose, to a \emph{quantizer}: a rule that maps a continuous input to the nearest of $k$ discrete representative values. 
Suppose such a quantizer partitions $[0,100]$ into $k$ equal-width bins, each of width $w = 100/k$, and reports the midpoint
of whichever bin the true input falls into. 
If the input $X \sim \text{Uniform}[0,100]$, then within any one bin, $X$ is uniformly distributed on an interval of width $w$ centered at that bin's midpoint. 
Writing $X$ relative to the midpoint as
$X' \sim \text{Uniform}[-w/2, w/2]$, the mean absolute error contributed by this bin is:

\begin{equation}
\mathbb{E}|X'| = \frac{1}{w}\int_{-w/2}^{w/2} |x|\,dx
= \frac{1}{w}\left[2\int_0^{w/2} x\,dx\right]
= \frac{1}{w}\cdot\frac{w^2}{4} = \frac{w}{4}.
\end{equation}
By symmetry, this is the same for every bin, so it is also the overall mean absolute error across the full range. Substituting $w = 100/k$:

\begin{equation}
\text{floor}(k) = \frac{100/k}{4} = \frac{25}{k}.
\end{equation}

This is the error a model would incur \emph{purely from having only
$k$ distinct values available}, even if it places those $k$ values optimally and reports them with perfect consistency. 
It therefore serves as a lower bound, not a prediction. 
A model's observed $\text{MARE}_m$ can equal $\text{floor}(k)$ only if its limited
vocabulary is used as efficiently as possible. 
The quantization floor represents the best theoretically achievable MARE for a k-level representation under a uniform input distribution. 
Excess error can result from either non-optimal placement/use of representational anchors or genuine mapping inconsistency.
\clearpage   

\end{document}